\documentclass[conference]{IEEEtran}
\IEEEoverridecommandlockouts
\usepackage{cite}
\usepackage{amsmath,amssymb,amsfonts}
\usepackage{algorithmic}
\usepackage{graphicx}
\usepackage{textcomp}
\usepackage{xcolor}
\def\BibTeX{{\rm B\kern-.05em{\sc i\kern-.025em b}\kern-.08em
    T\kern-.1667em\lower.7ex\hbox{E}\kern-.125emX}}
\begin{document}

\title{Self-Supervision, Remote Sensing and Abstraction: Representation Learning Across 3 Million Locations\\
}

\makeatletter
\newcommand{\linebreakand}{%
  \end{@IEEEauthorhalign}
  \hfill\mbox{}\par
  \mbox{}\hfill\begin{@IEEEauthorhalign}
}
\makeatother


\author{
    \IEEEauthorblockN{Sachith Seneviratne\IEEEauthorrefmark{1}, Kerry A. Nice\IEEEauthorrefmark{1}, Jasper S. Wijnands\IEEEauthorrefmark{1}, Mark Stevenson\IEEEauthorrefmark{2}\IEEEauthorrefmark{1}, Jason Thompson\IEEEauthorrefmark{1}}
    \IEEEauthorblockA{\IEEEauthorrefmark{1}Transport, Health, and Urban Design Research Lab,
Melbourne School of Design, University of Melbourne, Australia}
    \IEEEauthorblockA{\IEEEauthorrefmark{2}Melbourne School of Engineering, University of Melbourne, Australia
    \\\{sachith.seneviratne, kerry.nice, jasper.wijnands, mark.stevenson, jason.thompson\}@unimelb.edu.au}
}
\maketitle

\begin{abstract}
Self-supervision based deep learning classification approaches have received considerable attention in academic literature. However, the performance of such methods on remote sensing imagery domains remains under-explored. In this work, we explore contrastive representation learning methods on the task of imagery-based city classification, an important problem in urban computing. We use satellite and map imagery across 2 domains, 3 million locations and more than 1500 cities. We show that self-supervised methods can build a generalizable representation from as few as 200 cities, with representations achieving over 95\% accuracy in unseen cities with minimal additional training. We also find that the performance discrepancy of such methods, when compared to supervised methods, induced by the domain discrepancy between natural imagery and abstract imagery is significant for remote sensing imagery. We compare all analysis against existing supervised models from academic literature and open-source our models\footnote{https://github.com/sachith500/self-supervision-remote-sensing-abstraction} for broader usage and further criticism.
\end{abstract}


\section{Introduction}
Deep learning based methods have enabled significant breakthroughs in automating parts of the computer vision analysis process. However, these methods generally require a large amount of labelled data. Such data can be expensive and time consuming to collect, especially in domains such as remote sensing, where experts are required to annotate images. Reusing trained neural networks via transfer learning, which enables a model to be fine-tuned to perform a different task, has gained popularity due to the ability to address this problem to some extent. However, most neural networks are trained on tasks and imagery domains where image capture occurs from a horizontally captured perspective, whereas remote sensing operates on vertically captured imagery. Therefore, neural networks trained on datasets such as ImageNet, have limited utility for transfer learning in this regard. 

Due to the proliferation of satellite imagery programs such as Sentinel, unlabelled satellite imagery is readily available for researchers. Therefore, the ability to leverage unlabelled data for deep learning methods is desirable. Self-supervised learning methods\cite{chen2020simple,he2020momentum,oord2018representation} have emerged as a paradigm of deep learning capable of leveraging unlabelled imagery in such a manner. The advantage of these methods is in the ability for networks to learn in an unsupervised manner, while using a "pretext task" to train the network with a supervised loss function. Pretext tasks are such that labels and data required for the task can be automatically generated from unlabelled imagery data. The neural network is then trained on this pretext task, creating a representation usable in downstream tasks. This can include tasks such as jigsaw puzzle solving\cite{noroozi2016unsupervised}, relative position prediction\cite{doersch2015unsupervised} or image colorization\cite{zhang2016colorful}, which require the neural network to learn a generic representation of the dataset in order to solve the pretext task. While self-supervised methods have been exhaustively explored in horizontal perspective imagery domains, vertical imagery domains such as remote sensing imagery remain an under-explored area.

Remote sensing imagery, and vertical perspective images in general have some special considerations with respect to self-supervision. Computer vision analyses in these domains are generally conducted on datasets of a single resolution. This means that the observations under exploration tend to occur at similar sizes within the image, unlike in horizontal imagery, where the observation may be at various depths with respect to the image capturing sensor. Therefore, understanding how this affects self-supervised performance is important. Additionally, remote sensing imagery tends to have counterparts incorporating abstract information (such as in maps, a domain distinct from natural remote sensing imagery). Such information can also be utilized as imagery co-variates in analysis (such as in \cite{seneviratne2021contrastive}, which combines natural and abstract remote sensing imagery for species prediction) or form the basis for separate analysis. Such types of imagery include geographical information such as altitude levels (available as topographical maps such as from the NASA Shuttle Radar Topography Mission\footnote{https://lpdaac.usgs.gov/products/srtmgl1v003/}), land use data and general information available in maps (such Google maps, OpenStreetMaps\footnote{https://www.openstreetmap.org/}). This information is usually more abstract than natural images and presented using distinct colours for human consumption, introducing a clear domain discrepancy with respect to natural satellite imagery. Understanding how self-supervised methods operate on such data is important, as it affects how information from such sources can be used to augment remote sensing imagery analysis workflows, when using self-supervision. In general, exploration of self-supervised methods on imagery abstractions is poorly explored in academic literature, with most work focused on sketch based methods\cite{xie2019unsupervised,bhunia2021vectorization}.

Taking these factors into consideration, we explore three important questions in this paper. Firstly, given that remote sensing imagery may have low-level visual features in a more consistent manner than horizontal natural imagery, how does this affect the generalizability of representations learned in a self-supervised manner from unlabelled remote sensing images. We evaluate this by learning a representation on a small subset of classes from a remote sensing task, and evaluate the representation on the "hold-out" classes which the self-supervised representation has not been exposed to within the scope of a classification task. Secondly, we build representations for abstract (map) imagery which correspond to the vertically captured imagery domain due to the geographical information represented in them. We evaluate the performance of our representations and explore the question of how pretext task operations affect downstream task performance in abstract imagery. Finally, we quantitatively explore the domain discrepancy between using remote sensing imagery and imagery containing abstractions of remote sensing imagery in supervised and self-supervised workflows. We do this by comparing against existing supervised methods and results from the literature and by comparing the gap in performance between the supervised (existing work) and self-supervised (this paper) methods exploring the same problem. In order to evaluate the domain discrepancy effectively, we select a task that is achievable using both types of imagery and is identical in all respects. This allows for the exploration of model performance at several levels by evaluating: 

\begin{itemize}
    \item the supervised domain discrepancy between natural and abstract remote sensing imagery;
    \item the self-supervised domain discrepancy for the above; and
    \item the gap between the supervised and self-supervised domain discrepancies.
\end{itemize}

Given the above considerations, the choice of task to base our evaluation on is not straightforward. The task would require the following properties:
\begin{itemize}
    \item applicable to remote sensing (natural) imagery;
    \item applicable to abstractions of remote sensing imagery;
    \item availability of a large dataset covering multiple geographies; and
    \item is an established computer vision problem (for supervised comparison with existing literature).
\end{itemize}

A key challenge in this case is identifying a task that can be performed using both natural remote sensing imagery and abstracted imagery of vertical scenes. We find that the city prediction task defined in \cite{thompson2020global} meets all these requirements. In conjunction with this task, an imagery dataset spanning more than 3 million geographic locations was collected for robust evaluation. The task has been shown to have important implications in city design\cite{nice2019nature,nice2020paris}, health\cite{stevenson2021city} and injury prevention\cite{thompson2020global} due to the ability to allow the quantification of inter-city comparisons at a structural level. Most importantly, the task has more than 1500 classes per each of the two imagery domains, and 1000 images per class, which puts it at a scale roughly 3 times that of ImageNet\cite{krizhevsky2012imagenet}. Therefore, the city prediction task and associated datasets have been selected as the basis for analysis, while using RGB imagery during all stages of our analysis.

\section{Related Work}
\label{litreview}

\subsection{Optical remote sensing analysis using machine learning}

Remote sensing imagery analysis using machine learning techniques is a broad area in academic literature with much research attention dedicated to it. Most works explore natural remote sensing imagery\cite{dey2008decision,chen2018learning,wijnands2021identifying}, while some works explore abstractions of remote sensing imagery within the context of machine learning\cite{zhao2019unsupervised, nice2020paris,thompson2020global}. Methods in these areas almost exclusively follow supervised or unsupervised approaches, with very few, such as \cite{stojnic2021self}, following self-supervised approaches. \cite{seneviratne2021contrastive} and \cite{stojnic2021self} are of particular interest as they show that self-supervised pretraining (using Momentum Contrast\cite{chen2020improved} and Contrastive Multiview Coding\cite{tian2020contrastive} respectively) can outperform supervised pretraining in the remote sensing imagery domain, which forms the inspiration for most exploration in this paper. In this work, we explore established tasks from remote sensing based urban computing using machine learning techniques leveraging self-supervision, which falls under the broader area of deep learning.

\subsection{Self-supervised representation learning}
\label{lit:selfsupervision}
Self-supervised representation learning has received considerable attention in academic literature. These methods use a pretext task to perform representation learning. The pretext task forms a task that can usually be performed without any manual image annotation and thus allows supervised loss functions to be applied to unlabelled imagery. The selection of a specific self-supervised pretext task is an important consideration as it has considerable implications on the representation learned as well as the downstream tasks the representations are suitable for. MoCo\cite{he2020momentum}, SimCLR\cite{chen2020simple}, BYOL\cite{grill2020bootstrap}, SimSiam\cite{chen2020exploring} and InstDisc\cite{wu2018unsupervised} all use an instance discrimination pretext task. Methods such as SWAV\cite{caron2021unsupervised} operate on the task of cluster discrimination. Contrastive Predictive Coding\cite{oord2018representation} performs the pretext task of mutual information maximization. PIRL\cite{misra2020self} uses jigsaw puzzle solving, while RotNet\cite{komodakis2018unsupervised} uses rotation prediction. Several works such as BigGAN\cite{brock2019large} and BigBiGAN\cite{donahue2019large} draw insipiration from the image reconstruction problem commonly used in Autoencoders for performing self-supervised representation learning. However, the general performance of these methods on representation learning across different data domains is poorly understood\cite{cole2021does}. In this work we address two such domains: remote sensing and abstract imagery.

\subsection{Remote-sensing based self-supervision}
Very few works, such as \cite{maas2021seasonal,ayush2020geographyaware}, explore remote-sensing imagery-based self-supervision. However, the suitability of these methods for abstract imagery is uncertain. Most prior work leverages existing self-supervised workflows which have been extensively explored on varying datasets, such as those discussed in Section~\ref{lit:selfsupervision}. In this work we follow such work and leverage generic self-supervised methods for implementation on vertical perspective natural and abstract remote sensing imagery.

\subsection{Evaluation of Self-Supervision}
Evaluation of Self-Supervision is approached from several directions in prior research. In this work we use the standard frozen representation approach which is the commonly used protocol for evaluating the learned representation generated by a self-supervised method\cite{kolesnikov2019revisiting}. This involves evaluating the learned visual representation by using it for training a simple logistic regression model to solve a downstream multiclass image classification task. By freezing all parts of the network except this final linear layer, the network is unable to adapt to the task beyond combining the high-level features already encapsulated within the model. Thus, this allows objective criticism of the learned representation on the downstream task, while additionally minimizing computational requirements. This is because the number of trainable parameters in the network under this protocol is much lower compared to standard supervised training. Other approaches have also been proposed, such as employing Support Vector Machines (SVMs) on the output of the embedding learned by the model\cite{van2021benchmarking}. Multiple works benchmark self-supervised performance on different datasets\cite{goyal2019scaling,van2021benchmarking,kolesnikov2019revisiting} to enhance the understanding of method performance in different domains and downstream tasks. These works are particularly insightful due to the innately high computational requirements associated with building self-supervised models. Our work fits within this growing body of work as we compare supervised and self-supervised performance across an established classification task from academic literature. However, we additionally identify separate domains (and collect datasets) to apply the same task to, thus performing a stronger evaluation of the domain discrepancy and the gap in performance between self-supervised and supervised methods. 

In general, work discussed in the previous paragraph allows the discrimination of self-supervision based approaches. Such work is primarily driven by exploration of the performance gains within self-supervised methods. In principle, this area of research is largely similar to other work which explores feature extraction from deep neural networks such as \cite{shaheen2016impact}. Our work also falls into this category as it explores the performance of self-supervision on remote-sensing and abstract imagery, and contributes to the understanding of the behavior of representation learning techniques in the domains of remote sensing and abstract imagery. In particular, in this work we use self-supervision techniques and the ResNet50\cite{he2016deep} and Vision Transformer\cite{dosovitskiy2020image} (ViT) architectures as deep feature extractors.

\subsection{Self-supervision and abstract imagery}
The intersection of self-supervision and abstract imagery is an underexplored area in academic literature. Most prior work\cite{xie2019unsupervised,bhunia2021vectorization} focuses on sketch based abstractions. We hypothesize that this is because the utility of image abstractions is limited outside of the remote sensing domain. This is because for the horizontal perspective, natural images have too much variation within them, as there is considerable spatial (depth of object of interest) variation between images. In remote sensing imagery, this variation (we hypothesize) is lower due to the fixed perspective of the camera and the fixed resolution of the imagery. Additionally, abstract imagery in remote sensing provides more useful "knowledge" for human consumption. In horizontal perspective natural imagery, such information is usually generated by semantic segmentation or object detection techniques, which have error rates associated with them. This limits the utility of their use as co-variates for analysis in data fusion approaches. This work contributes to the understanding of the performance of self-supervised methods on abstract imagery, with a particular focus on abstractions of remote sensing imagery. In particular, the evaluation of learnt representations on down-stream tasks provides contributions to their use in ensembling approaches, either using self-supervised representations (such as in \cite{seneviratne2021contrastive}) or otherwise.

\section{Methodology}

\subsection{Datasets}
\label{dataset}

Imagery from the largest global cities with populations greater than 300,000 people, as specified by \cite{UN2014}, was collected from Google Maps and has been used in a number of studies\cite{nice2019nature,nice2020paris,thompson2020global} examining urban form and to quantify the structural similarity between cities. The sampling procedure used a circular sampling area with the radius scaled by population size by a power of 0.85, according to \cite{Barthelemy2016}, and aligned to the city's centre (as specified by \cite{UN2014}) to provide a standardised method to define urban boundaries. Large water-bodies were removed from the sampling area. These procedures resulted in a population and water body-adjusted circular area centred on the city's central coordinates, capturing the widest extent of each city while minimising the amount of non-urban locations. 

Random sampling of 1000 images within these boundaries was performed for each city to build two types of imagery datasets. The first was of map imagery, using the Google Static Maps API\cite{GoogleStatic2017} with an image type of `map'. Images were sized 256$\times$256 pixels using a zoom level of 16 (approximately 400m$\times$400m). These were obtained from the selected locations using a custom style. This style abstracted roads, public transport networks, green space, and water bodies into solid colors of black, orange, green, and blue, respectively. Any remaining space was coded white. Due to mapping inconsistencies in South Korea, all 25 South Korean cities were removed from the dataset, reducing the number of cities to 1665 for a total data set of 1,665,000 images in 1665 classes (i.e., the corresponding city names). See Figure~\ref{fig:maps}a for an example, of Paris, France.

A second set of imagery was collected using the Google Static Maps API \cite{GoogleStatic2017} with an image type of `satellite', zoom level of 16 (approximately 400$\times$400m), and image size of 256$\times$256. Google Maps satellite imagery is a mosaic of cloud-free imagery from multiple sources and with different acquisition times. Originally based on Landsat 7 imagery, this has largely been replaced by Landsat 8 since 2013 and has a 15m/pixel resolution \cite{Google2013}. The majority of the imagery collected dated from March and April 2017, with the rest from 2016 and early 2017. A few locations (such as Iraq and Afghanistan) date back as far as 2010. Suitable imagery was not available for two cities, bringing the number of cities to 1690 and leading to a total data set of 1,690,000 images in 1690 classes. Figure~\ref{fig:maps}b shows a sample image, from Adelaide, Australia. A list of all locations used in this dataset is available \footnote{https://github.com/sachith500/self-supervision-remote-sensing-abstraction}, so that our results are fully reproducible.


\begin{figure}[!htbp]
 \centering
a)\includegraphics[scale=1.22]{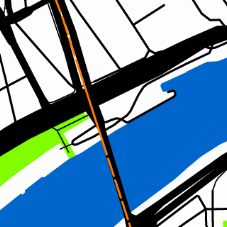} 
b)\includegraphics[scale=0.4]{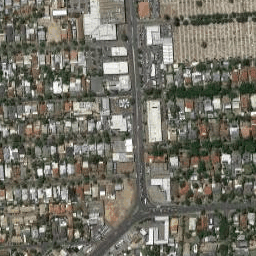}
\caption{\bf Sample imagery for a) Google Maps dataset (from Paris, France \cite{GoogleStatic2017}) and b) Google Satellite dataset (from Adelaide, Australia \cite{GoogleStatic2017}).
} 
 \label{fig:maps}
\end{figure}

A random train-test split of 80\% vs 20\% was used for analysis, providing 800 training images and 200 testing images per class (per city). The same data splits are used in all experiments. Representation learning is performed only on the training split.

\subsection{Task Definition}
\label{taskdef}
The task we explore is a classification task, with many similarities to ImageNet classification\cite{krizhevsky2012imagenet}, but from a remote sensing/vertical perspective. There are more than 1500 classes of images per dataset, with each image belonging to a particular city. The task we explore is the classification of those images to the city it was captured in. To accomplish this, the neural network must build a set of features to identify the unique characteristics of each city. This is generally quite difficult for humans to do without unique identifiable landmarks present, however neural networks have been shown to achieve a much higher accuracy, even over a large number of classes\cite{nice2020paris}. This task is ideal for evaluation of class-independent representation learning capabilities, as the intra-class differences are low. This is because each image is from a vertical perspective of a city and would contain similar low-level image features such as trees, buildings and roads. Additionally, the possibility of performing the same task on both natural and abstract imagery allows us to better explore the domain discrepancy between the types of imagery across both supervised and self-supervised methods.

\subsection{Self-supervised Representation Learning}
\label{selfsup}

Self-supervised learning utilizes pretext tasks to train a model in a supervised manner. Contrastive Representation Learning is a subset of self-supervised representation learning which exploits the similarity between augmented views of the same image, and the dissimilarity between different images, using a contrastive loss such as InfoNCE\cite{oord2018representation}. This contrastive loss is computed based on the query's ability to match positive samples corresponding to augmentated views of the source image together, with negative samples corresponding to augmented views of other images acting to confound this classification. Here, the query denotes an augmented view of the source image. InfoNCE can be considered as a softmax loss with the objective of classifying the query as belonging to the class corresponding to the positive sample (with the other negative samples acting as negative classes).

We utilize three such representation learning workflows in this work, which are annotated as \textbf{V1}, \textbf{V2} and \textbf{DINO} for easier comparison. We utilize the ResNet50 and ViT-S/16 architectures in our evaluation.

\textbf{V1: Momentum Contrast}

The first workflow, annotated as \textbf{V1}, used for representation learning utilizes Momentum Contrast (MoCo)\cite{he2020momentum}, which utilizes a queue and a momentum update in order to avoid the need to use large memory banks corresponding to encoded representations of the dataset. The pretext task used is instance discrimination \cite{wu2018unsupervised} and we set experimental settings to be consistent with \textbf{V2}. We use the parameters suggested in \cite{he2020momentum} -- learning rate of 0.03 and a batch size of 256 on a 4 GPU node.

\textbf{V2: V1 with projection head and augmentations}

The second workflow utilizes the above workflow, but additionally incorporates aspects of SimCLR\cite{chen2020simple} into the representation learning workflow in the form of additional augmentations and a projection head. This workflow has been shown to significantly improve performance of the learned represesentations on downstream tasks\cite{chen2020improved}. In our experiments, we evaluate \textbf{V1} and \textbf{V2} in order to better understand performance across both natural and abstract vertical perspective imagery. Of particular interest is the effect of the extra augmentations in \textbf{V2} (some of which are based on distorting the colours in the image) on the performance of representations on downstream abstract imagery (which use colour to indicate important aspects of the image). We use the same parameters as \textbf{V1} for consistency - with a learning rate of 0.03, a batch size of 256 on a 4 GPU node, a softmax temperature of 0.2, and the additional augmentations and cosine learning rate schedules suggested in \cite{chen2020improved}.

\textbf{DINO: Self Distillation with Vision Transformers}

DINO\cite{caron2021emerging} utilizes representation learning through knowledge distillation to perform self-supervised learning. Student and teacher networks are trained on different zoomed views of the data. The student network generally encodes smaller crops of the image, while the teacher network encodes larger crops of the image. The encodings generated by both networks are passed through a softmax function effectively mapping the encodings to a probability distribution over a large number of psuedo-classes (65,536 by default and in our work). By calculating the loss as a cross-entropy function over these psuedo-classes, the student network is updated. The teacher network is only modified using a momentum based exponential moving average update based on the student parameters. 

We evaluate the ViT-S/16 architecture using this workflow as the basis for representation learning. This architecture has 21 million parameters (comparable to ResNet50 with 23 million). We distribute pre-training on 28 GPUs over a week, using a patch-size of 16 to enable faster training. Otherwise, we use the same hyper-parameters as those suggested for ImageNet in \cite{caron2021emerging}. Due to the higher computational requirements in this workflow, we only evaluate it on abstract imagery.

\begin{table*}[htbp]
\centering
 \begin{tabular}{||c|c|c|c|c|c|c|c||} 
 \hline
 Row & Experiment & Imagery & Workflow & Pretrain cities& Pretrain epochs & test cities&  Accuracy \\ [0.5ex] 
 \hline\hline
 1 & Generalizability & Satellite& MoCo V1 & 200 & 200 & 200 & 95\% \\ 
 \hline
2 & Generalizability & Satellite& MoCo V2 & 200 & 200 & 200 & 99\% \\ 
 \hline
 3 & Generalizability & Satellite& MoCo V1 & 200 & 200 & 1690 & 81\% \\ 
 \hline
 4 & Generalizability & Satellite& MoCo V2 & 200 & 200 & 1690 & 95\% \\ 
 \hline
 
 5 & Generalizability & Satellite& MoCo V2 & 1690 & 145$^{\mathrm{1}}$  & 1690 & 98\% \\ 
 \hline
 6 & Abstraction & Maps& MoCo V1 &1665 & 200 & 1665 & 67\% \\
 \hline
 7 & Abstraction & Maps& MoCo V2 &1665 & 200 & 1665 & 61\% \\
 \hline
 8 & Abstraction & Maps& DINO &1665 & 180$^{\mathrm{1}} $& 1665 & 36\% \\
 \hline
\multicolumn{4}{l}{$^{\mathrm{1}}$terminated at the end of 1 week of wall clock time.}
\end{tabular}
 \caption{Top-1 Accuracy Results for generalizability and abstraction experiments (from Sections~\ref{experiment:generalizability} and~\ref{experiment:abstraction}). }
\label{tab:generalizability}
\end{table*}
\subsection{Evaluation}

Evaluation of the workflows from Section~\ref{selfsup} proceeds according to the standard protocol for representation learning, as discussed in Section~\ref{litreview}. First, \textbf{V1} or \textbf{V2} is used to build a representation with a ResNet50 architecture. Then, a fully connected linear layer is built following the final bottleneck layer of the ResNet50. This layer connects the outputs of the bottleneck layer (with 2048 neurons) to the class prediction output neurons (1665 or 1690 neurons depending on imagery type and equal to the number of classes in each task). This single layer is trained on the task, leveraging the representation from \textbf{V1} or \textbf{V2} using a learning rate of 30 and a batch size of 256 on a 4 GPU node for a total of 100 epochs with learning rate decreased by a factor of 10 at the 60th and 80th epochs, respectively. Categorical cross-entropy was used as the loss function. The large learning rate is justified as most of the network is frozen\cite{he2020momentum}.

For the evaluation using Vision Transformers, we use an approach similar to the above. The ViT-S/16 architecture contains 12 Transformer blocks, from which we use the last 4 and concatenate the output. As the output of each block has a dimension of 384, the combined output is 1536 dimensional. This combined output is connected to the class prediction output neurons (1665 in this case) in a fully connected manner.


\subsection{Experiment 1: Representation Generalizability}
\label{experiment:generalizability}
In this experiment, we evaluate the class generalizability of the representation learned using the methodology defined in Section~\ref{selfsup}, evaluating workflows (\textbf{V1} and \textbf{V2}). This is done by evaluating a representation learned on a subset of cities on the city prediction task defined in Section~\ref{taskdef}. In this experiment we use the satellite imagery dataset described in Section~\ref{dataset}. Representation learning is performed on images from a subset of available cities, in an unsupervised manner with different experimental variations based on workflow, number of cities for representation learning and number of cities for evaluation. The cities used for representation learning are always used for evaluation, with additional holdout cities included as well. For robust evaluation, evaluations were carried out on the 200 cities used for representation learning, and separately on the full 1665/1690 cities as well. Evaluation was performed by first training a new linear layer on top of a given representation as described in Section~\ref{selfsup}. The representation has its weights frozen which forces the new linear layer to use the provided unsupervised representation without any further modification. This is followed by evaluation on the test set. We report the best performance across the different experimental settings in Section~\ref{results}. We allow the representation learning process a maximum of either one week (wall-clock time) of training on a 4 GPU node or 200 epochs of training. Where this led to termination of the process before 200 epochs, we use the representation generated in the final epoch prior to termination. This is in line with prior work\cite{chen2020improved} which indicates that longer pretraining on the pretext task provides better representations as evidenced by evaluation on downstream tasks.

\subsection{Experiment 2: Abstraction}
\label{experiment:abstraction}
This experiment is identical to the previous experiment in all respects, except we evaluate on the additional domain of abstract map imagery (across 1665 cities as discussed in Section~\ref{dataset}). In this experiment, we extend representation learning time to two weeks on a 4 GPU node, allowing all experiments to reach 200 epochs of training prior to termination. This is to keep the representation learning process consistent allowing for better comparability of results as we always pick the latest representation generated by the self-supervised training on the pretext task. We present results in a manner consistent with the previous experiments for easier comparability within the same table (Table~\ref{tab:generalizability}).

\section{Results}
\label{results}
\subsection{Generalizability and Abstraction}

The combined results of the generalizability experiment defined in Section~\ref{experiment:generalizability} and the abstraction experiment in Section~\ref{experiment:abstraction} are presented in Table~\ref{tab:generalizability}. This table contains several interesting results generated by different representations trained with varying data using self-supervised representation learning. "Pretrain cities" indicate the number of cities used for representation learning. "Test cities" indicate the number of cities used for evaluation, and is equal to the number of cities seen by the representation ("Pretrain cities") and the number of cities unseen by the representation. It is important to note that the representations remain frozen during the training process, which means they are unable to perform any additional learning on cities during this stage. Thus, the representation is trained only on cities listed as "Pretrain cities" in the results, with other cities remaining unseen. This highlights the generalizability of the methods presented in this work.

Based on this table, we can make the following observations and draw several conclusions:

\begin{itemize}
    \item Workflow \textbf{V2} is superior to \textbf{V1} for natural satellite imagery (Rows 1 and 2, Rows 3 and 4). This discrepancy is more pronounced when evaluating on classes unseen by the representation (Row 3 and 4 - 14\% discrepancy in accuracy). This implies that workflow \textbf{V2} is able to learn more generalizable representations with respect to unseen classes. This means that workflows using \textbf{V2} are able to learn a more general representation from classes used for representation learning, and able to maintain performance on classes unseen during the representation learning process compared to \textbf{V1}.
    \item The representations learned using \textbf{V2} on natural imagery across 200 classes are able to generalize well over 1490 unseen classes (Rows 4 and 5 indicate only a 3\% discrepancy between pretraining on 200 vs 1690 classes).
    \item For abstract map imagery, \textbf{V2} underperforms \textbf{V1} (Rows 6 and 7). We suspect this is due to the importance of colours in the abstract imagery classification task, however further studies are required before any conclusive general conclusions can be drawn.
    \item Results indicate a significant domain discrepancy for self-supervised methods between abstract and natural imagery in remote sensing (Rows 5 and 6).
\end{itemize}

Importantly, by limiting representation learning to 200 cities (Row 1) as part of this experiment, we show that the self-supervised workflow is able to learn generic features that makes it generalizable across 1490 unseen cities. This provides confidence that the method we utilize is able to learn a generic representation capable of generalizing across geographies, and learn discriminatory features that are useful across many unseen classes (cities).

\subsection{Domain Discrepancy and Supervised Comparison}

In this section we compare the results generated in our experiments against existing results from the literature. While we use a different backbone for our analysis, the similar results (98\% vs 99\% in prior work) generated by our method on satellite imagery indicates that this architecture is suited to the task. These comparable results on the natural imagery city classification task allow us to directly compare the results generated by the other self-supervised training configurations we explore in this work, with results from prior work in the literature.

\begin{table}[htbp]
 \begin{tabular}{||c|c|c|c|c|c||} 
 \hline
 Work & Imagery & Method & Architecture & Test cities&  Acc \\ [0.5ex] 
 \hline\hline
 \cite{nice2020paris}& Satellite & Supervised & InceptionV3 & 1690 & 99\% \\ 
 \hline
 \cite{thompson2020global}& Maps & Supervised & InceptionV3 & 1665 & 86\% \\ 
 \hline
 Ours& Satellite & Self-supervised & ResNet50 & 1690 & 98\% \\ 
 \hline
 Ours& Maps & Self-supervised & ResNet50 & 1665 & 67\% \\ 
 \hline
\end{tabular}
 \caption{Comparison with existing work in the literature}
 \label{tab:comparewithlit}
\end{table}

Table~\ref{tab:comparewithlit} indicates that self-supervised methods are capable of generating similar results for this task on satellite imagery, but indicates a large discrepancy for map imagery. This is further highlighted in Table~\ref{tab:domaingap}, which indicates a 31\% domain discrepancy in performance between satellite and map imagery while only a 13\% gap is observed under supervised settings. Due to performing evaluation under identical experimental settings, we can draw the following conclusions based on these results:

\begin{itemize}
    \item Self-supervised methods are \textbf{able} to achieve comparable performance to supervised methods for \textbf{natural} imagery on this task (Column "Satellite" of Table~\ref{tab:domaingap}).
    \item Self-supervised methods display a significant domain discrepancy compared to supervised methods for \textbf{abstract} imagery on this task (Column "Map" of Table~\ref{tab:domaingap}). Clearly, the domain discrepancy is more significant in self-supervision under identical evaluation settings. 
\end{itemize}

\begin{table}[htbp]
 \begin{tabular}{||c|c|c|c||} 
 \hline
 Method & Satellite & Map & Difference (Satellite - Map) \\ [0.5ex] 
 \hline\hline
 Supervised &  99\% & 86\% & 13\%  \\ 
 \hline
 Self-supervised & 98\% & 67\% & 31\% \\ 
 \hline
 
\end{tabular}
 \caption{Domain discrepancy - supervised vs self-supervised}
\label{tab:domaingap}
\end{table}

\section*{Conclusion}

In this paper we show that self-supervised methods are able to achieve similar performance to supervised methods on natural satellite imagery. We base our analysis on a well motivated task from the literature on more than 3 million images across 2 domains with over 1500 classes. We show that remote sensing imagery representations built using natural imagery from as few as 200 classes can be effectively expanded to more than 1000 unseen classes with minimal additional training to achieve comparable performance to supervised workflows. This is important as the self-supervised representations may be utilized in different tasks with minimal additional training.  We show that the domain discrepancy introduced in abstractions of imagery based on remote sensing data is significant. We find that this domain discrepancy is further exaggerated in self-supervised workflows. We share our trained models\footnote{Trained models available at: https://github.com/sachith500/self-supervision-remote-sensing-abstraction} and inference scripts, with a list of exact locations used in our study so that our results are fully reproducible. Overall, our work indicates considerable promise for self-supervision based methods for the area of remote sensing. However, our experimental results indicate that the ability for these methods to operate well on natural imagery does not translate well for handling abstract imagery. In particular, our comparisons with past work indicate that supervised neural networks are able to learn compatible representations on the task of abstract imagery classification. Therefore, we believe that the large discrepancy we observe in the performance of state of the art self-supervision techniques in this domain warrants further investigation.

\section*{Acknowledgments}

This project is supported by National Health and Medical Research Council Grant GA80134. This research was undertaken using the LIEF HPC-GPGPU Facility hosted at the University of Melbourne. This Facility was established with the assistance of LIEF Grant LE170100200. This research was undertaken using University of Melbourne Research Computing facilities established by the Petascale Campus Initiative.

{ \small
\bibliography{references3}
\bibliographystyle{plain}
}


\end{document}